# A photosensor employing data-driven binning for ultrafast image recognition

Lukas Mennel[1], Aday J. Molina-Mendoza[1], Matthias Paur[1], Dmitry K. Polyushkin[1], Dohyun Kwak[1], Miriam Giparakis[2], Maximilian Beiser[2], Aaron Maxwell Andrews[2], and Thomas Mueller[1‡]

[1] *Vienna University of Technology, Institute of Photonics, Gußhausstraße 27-29, 1040 Vienna, Austria*
[2] *Vienna University of Technology, Institute of Solid-State Electronics, Gußhausstraße 25, 1040 Vienna, Austria*

‡ Corresponding author: thomas.mueller@tuwien.ac.at

**Pixel binning is a technique, widely used in optical image acquisition and spectroscopy, in which adjacent detector elements of an image sensor are combined into larger pixels. This reduces the amount of data to be processed as well as the impact of noise, but comes at the cost of a loss of information. Here, we push the concept of binning to its limit by combining a large fraction of the sensor elements into a single "superpixel" that extends over the whole face of the chip. For a given pattern recognition task, its optimal shape is determined from training data using a machine learning algorithm. We demonstrate the classification of optically projected images from the MNIST dataset on a nanosecond timescale, with enhanced sensitivity and without loss of classification accuracy. Our concept is not limited to imaging alone but can also be applied in optical spectroscopy or other sensing applications.**

With the recent advances in machine vision applications, there is a growing demand for sensor hardware that is faster, more energy-efficient, and more sensitive than frame-based cameras such as charge-coupled devices (CCDs) or complementary metal-oxide-semiconductor (CMOS) imagers[1,2]. Beyond event-based cameras (silicon retinas)[3,4], which rely on conventional CMOS technology and have reached a high level of maturity, there is now increasing research on novel types of image acquisition and data pre-processing techniques[5,6,7,8,9,10,11,12,13,14,15,16,17], with many of them emulating certain neuro-biological functions of the human visual system.

One image pre-processing technique, that is being used since decades, is pixel binning. Binning is the process of combining the electric signals from $K$ adjacent detector elements into one larger pixel. This offers benefits such as (i) increased frame rate due to a $K$-fold reduction in the amount of output data, and (ii) an up to $K^{1/2}$-fold improvement in signal-to-noise ratio (SNR) at low light levels or short exposure times [18]. The latter can be understood from the fact that dark noise is collected in normal mode for every detector element, but in binned mode only once per $K$ elements. Binning, however, comes at the expense of reduced spatial resolution or, in more general terms, loss of information. In pattern recognition applications this reduces the accuracy of the results even if the SNR is high (Supplementary Figure S1).

In Figure 1a we schematically depict different types of binning. Besides the aforementioned conventional approach, we also illustrate our concept of data-driven binning. There, a substantial fraction of pixels are combined into a "superpixel" that extends over the whole face of the chip, thus forming a large-area photodetector with a complex geometrical structure that is determined from training data. For multi-class classification with one-hot encoding, one such superpixel is required for each class. As for conventional binning, the system becomes more resilient towards noise and its dynamic range increases. However, there is no loss of information and hence no compromise in performance. These benefits come at the cost of less flexibility, as a custom configuration/design is required for each specific application.

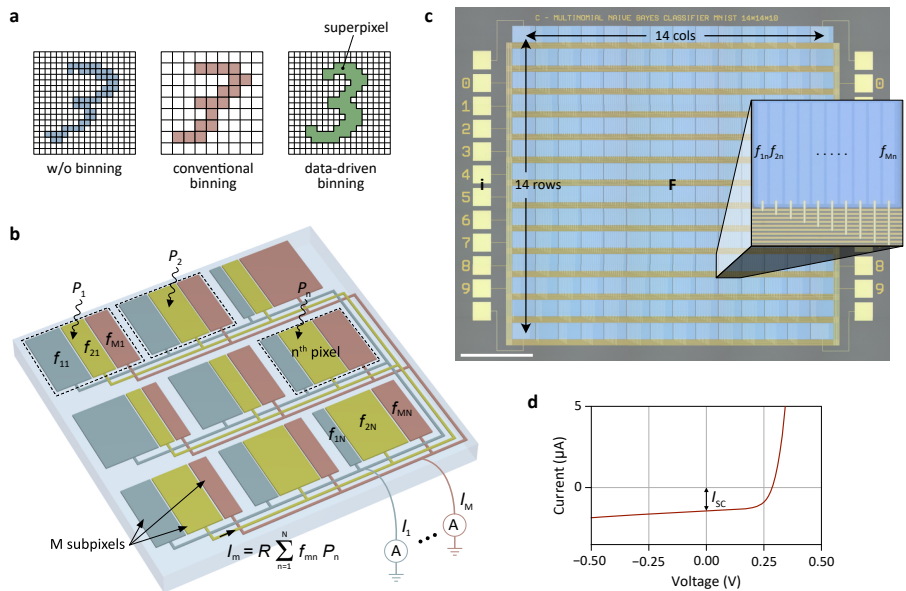

**Figure 1 | Photosensor implementation. a,** Different types of pixel binning. **b,** Schematic illustration of the photosensor. Each pixel is divided into subpixels that are connected together to form $M$ superpixels, whose output currents $I_m$ are measured. **c,** Microscope image of a NB classifier for

MNIST classification with $N = 14 \times 14$ pixels and $M = 10$ output channels. Scale bar, 500 μm. Inset: microscope image of the $n$-th pixel showing $M$ subpixels with different fill factors $f_{mn}$. **d,** Current-voltage characteristic for one of the detector elements under optical illumination. $I_{SC}$ is the short-circuit photocurrent.

Figure 1b shows a schematic of our photosensor. A microscope photograph of the actual device implementation is shown in Figure 1c. For details regarding the fabrication, we refer to the Methods section. The device consists of $N$ pixels, arranged in a two-dimensional array. Each pixel is divided into at most $M$ subpixels that are connected—binned—together to form the $M$ superpixels, whose output currents are measured. Each detector element is composed of a GaAs Schottky photodiode (Supplementary Figure S2) that is operated under short-circuit conditions (Figure 1d) and exhibits a photoresponsivity of $R = I_{SC}/P \approx 0.1$ A/W, where $I_{SC}$ is the photocurrent and $P$ the incident optical power. GaAs was chosen because of its short absorption and diffusion lengths, which both reduce undesired cross-talk between adjacent pixels; with some minor modifications the sensor can also be realized using Si instead of GaAs. The design parameters, that depend on the specific classification task and are determined from training data, are the geometrical fill factors $f_{mn} = A_{mn}/A$ for each of the subpixels, where $A_{mn}$ denotes the subpixel area and $A$ is the total area of each pixel. From Figure 1b, we find for the $m$ output currents $I_m = R \sum_{n=1}^{N} f_{mn} P_n$, or

$$\mathbf{i} = R\mathbf{F}\mathbf{p}, \quad (1)$$

with $\mathbf{p} = (P_1, P_2, \dots, P_N)^T$ being a vector that represents the optical image projected onto the chip, $\mathbf{i} = (I_1, I_2, \dots, I_M)^T$ the output current vector, and $\mathbf{F} = (f_{mn})_{M \times N}$ a fill factor matrix that depends on the specific application. The $m$-th row of $\mathbf{F}$ is a vector $\mathbf{f}_m = (f_{m1}, f_{m2}, \dots, f_{mN})^T$ that represents the geometrical shape of the $m$-th superpixel.

Let us now discuss how to design the fill factor matrix for a specific image recognition problem. As an instructive example, we present the classification of handwritten digits ('0', '1', …, '9') from the MNIST dataset[19] by evaluating the posterior $\mathbb{P}(y_m|\mathbf{p})$ (the probability $\mathbb{P}$ of an image $\mathbf{p}$ being a particular digit $y_m$) for all classes and selecting the most probable outcome. By applying Bayes' theorem and further assuming that the features (pixels) are conditionally independent, one can derive a predictor of the form $\hat{y}_m = \arg\max_{m \in \{1\dots M\}} \mathbb{P}(y_m) \prod_{n=1}^{N} \mathbb{P}(P_n|y_m)$, known as Naïve Bayes (NB) classifier[20,21]. We use a multinomial event model $\mathbb{P}(P_n|y_m) = \pi_{mn}^{P_n}$, where $\pi_{mn}$ is the probability that the $n$-th pixel for a given class $y_m$ exhibits a certain brightness, and express the result in log-space to obtain a linear discriminant function

$$\hat{y}_m = \underset{m \in \{1...M\}}{\arg\max}\, (\mathbf{W}\mathbf{p} + \mathbf{b})_m \qquad (2)$$

with weights $w_{mn} = \log \pi_{mn}$. The bias terms $b_m = \log \mathbb{P}(y_m)$ can be omitted ($\mathbf{b} = 0$), as all classes are equiprobable. The similarity to Eq. (1) allows us to map the algorithm onto our device architecture: $\mathbf{F} \propto \mathbf{W}$. In order to match the calculated $w_{mn}$-value range to the physical constraints of the hardware implementation,

$$0 \leq f_{mn} \leq 1 \quad \text{and} \quad \sum_m f_{mn} \leq 1, \qquad (3)$$

we normalize the weights according to

$$f_{mn} = \frac{w_{mn} - \min\, w_{mn}}{\max\, \sum_m (w_{mn} - \min\, w_{mn})}. \qquad (4)$$

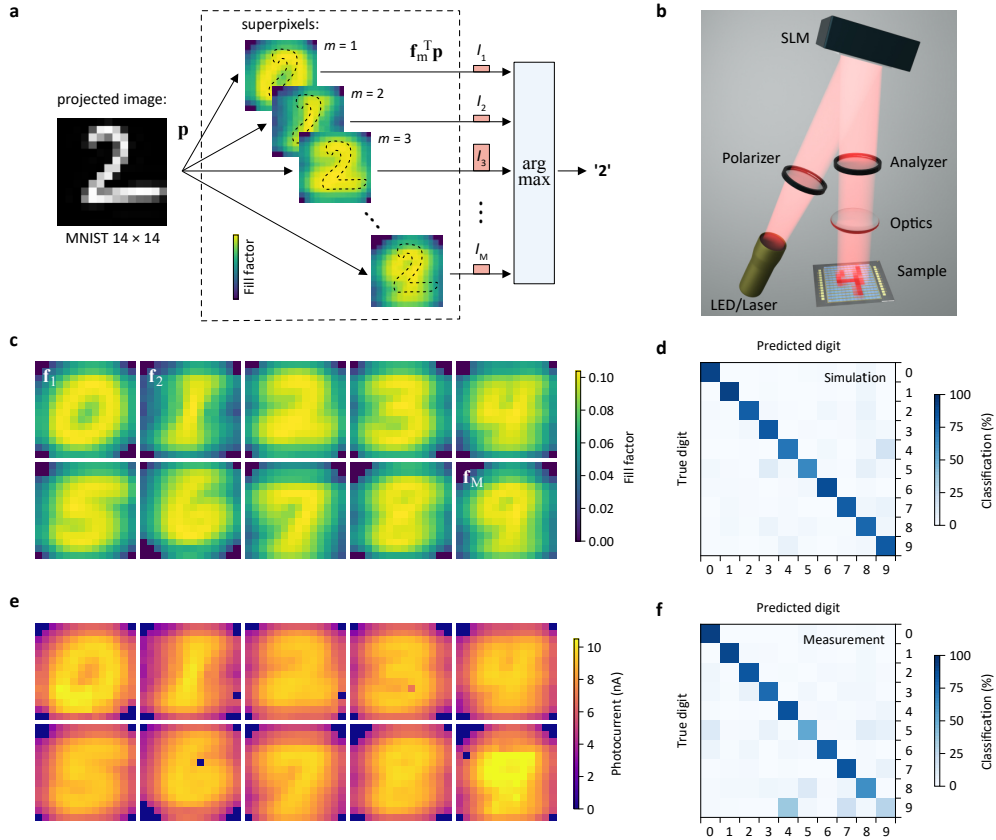

**Figure 2 | Naïve Bayes photosensor. a,** Schematic illustration of the working principle. An image from the MNIST dataset is projected onto the chip and detected by each superpixel. The channel with the largest output current is selected. We perform this operation in the digital domain; in the analogue it could be realized by a winner-take-all circuit[22]. **b,** Sketch of the experimental setup. **c,** Superpixel shapes for the NB classifier as determined from the MNIST training dataset. **d,** Calculated confusion matrix. **e,** Measured photoresponsivity maps. **f,** Experimental confusion matrix.

In Figure 2a we exemplify the working principle of the photosensor. A sample $\mathbf{p}$ from the MNIST dataset is optically projected onto the chip using the measurement setup shown in Figure 2b (see Methods section for experimental details). Each of the $M$ superpixels generates a photocurrent $I_m$ proportional to the inner product $\mathbf{f}_m^T\mathbf{p}$. If we visualize $\mathbf{f}_m$ for each class (Figure 2c), we obtain an intuitive result: The shape of each superpixel resembles that of the "average-looking" digit for the respective class. It is apparent that the superpixel with the largest spatial overlap with the image delivers the highest photocurrent.

Figure 2e shows experimental photocurrent maps for the device in Figure 1c. Here, each pixel of the sensor is illuminated individually and the output currents are recorded. The currents are proportional to the designed fill factors in Figure 2c, confirming negligible cross-talk between neighbouring subpixels. To evaluate the performance, we project all $10^4$ digits from the MNIST test dataset and record the sensor's predictions. The classification results are presented as a confusion matrix in Figure 2f. The chip is able to classify digits with an accuracy that closely matches the theoretical result in Figure 2d.

Beyond the instructive example of NB, the same device structure also allows the implementation of other, more accurate, classifiers. Specifically, we present the realization of a single-layer ANN[20] for the same MNIST classification task as discussed before. In Figure 3a the architecture of the network is shown. It makes its predictions according to

$$\hat{y}_m = \underset{m\in\{1...M\}}{\arg\max}\ \sigma(\mathbf{W}\mathbf{p} + \mathbf{b})_m \quad (5)$$

Note the similarity to Eq. (2), apart from a nonlinearity $\sigma$ which can be readily implemented, either in the analogue or the digital domain, using external electronics. We choose a softmax activation function for $\sigma$. Again, due to the physical constraints of the sensor hardware, we train the network with bias $\mathbf{b} = 0$ using categorical cross-entropy loss. In order to obey Eq. (3), we further introduce a constraint that enforces a non-negative weight matrix $\mathbf{W}$ by performing the following regularization after each training step:

$$\mathbf{W} \leftarrow \mathbf{W} \odot \theta(\mathbf{W}), \quad (6)$$

with $\odot$ denoting the Hadamard product and $\theta$ the Heaviside step function. This leads to a <1% penalty in accuracy.

The fill factor matrix $\mathbf{F}$, plotted in Figure 3d, is directly related to $\mathbf{W}$ by a geometrical scaling factor. Although the superpixel shapes do not clearly resemble the handwritten digits, the ANN shows better performance than the NB classifier, as demonstrated by the confusion matrix in Figure 3b. In addition, the ANN shows a larger spread between the highest and all other output currents (Figure 3c), which makes it more robust against noise (Supplementary Figure S3).

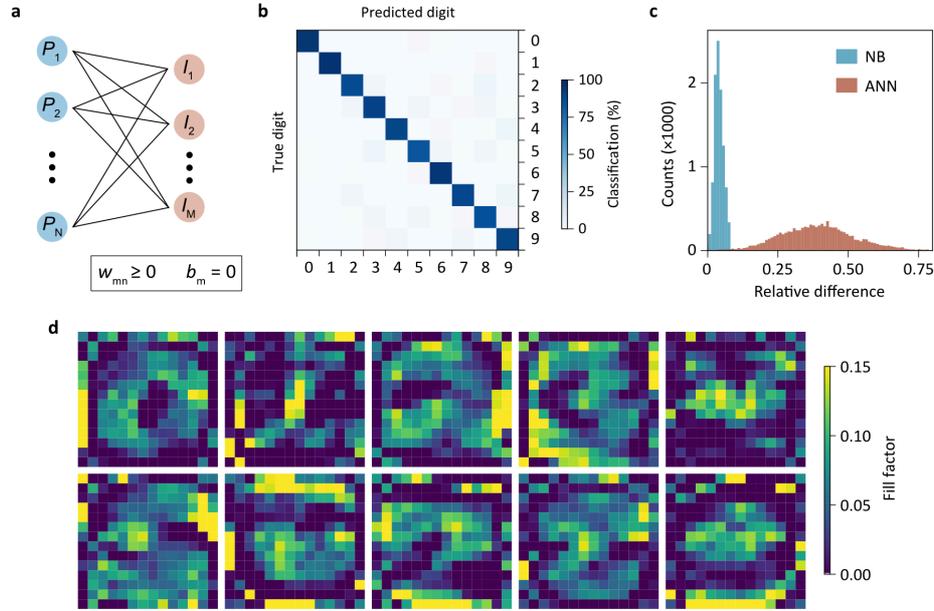

**Figure 3 | ANN photosensor. a**, Sketch of the ANN with weight and bias constraints. **b,** Confusion matrix for the ANN sensor. **c**, Relative difference between the highest and all other output currents. The ANN exhibits a larger spread in output currents than the NB classifier. **d**, Superpixel shapes for the ANN.

In Figure 4 we demonstrate the benefits of data-driven binning. It is evident that the readout of $M$ photodetector signals requires less time, resources, and energy than the readout of the whole image in a conventional image sensor. In fact, the photodiode array itself does not consume any energy at all; energy is only consumed by the electronic circuit that selects the highest photocurrent. Pattern recognition and classification occur in real-time and are only limited by the physics of the photocurrent generation and/or the electrical bandwidth of the data acquisition system. This is demonstrated in Figure 4a, where we show the correct classification of an image on a nanosecond timescale, limited by the bandwidth of the used amplifier.

Furthermore, it is known that binning can offer an $K^{1/2}$-fold improvement in SNR[18]. In our case, a substantial fraction $\xi$ (~60% for NB) of all sensor pixels are binned together ($K = \xi N$), with each pixel being split into $M$ elements. Together, this results in a $(\xi N)^{1/2}/M$-fold

SNR gain over the unbinned case. In Figure 4b we present experimental results for a binary classification task ('0' versus '1') at different illumination intensities. The classification accuracy is affected by the amplifier noise. For large intensities, the system operates with its designed accuracy. As the intensity is decreased, the classification accuracy drops and eventually, when the noise dominates over the signal, reaches the baseline of random guessing. Our device, employing data-driven binning, can perform this task at lower light intensities than the reference device without binning (see Methods section for details).

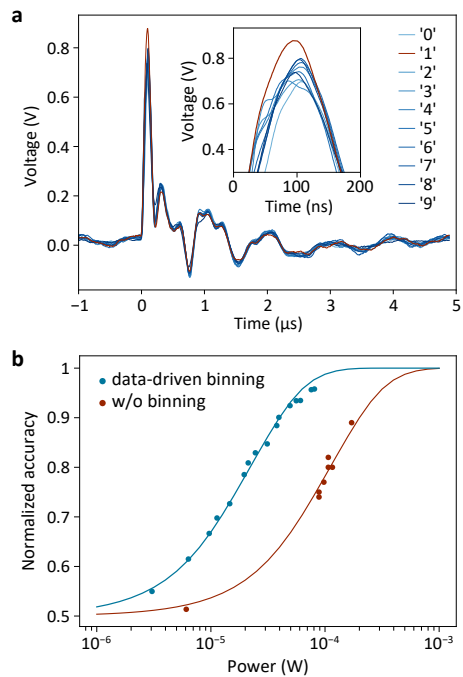

**Figure 4 | Evaluation of the device performance. a**, Demonstration of the high-speed capabilities of the sensor, measured with a 40-ns pulsed laser source. A '1' is projected onto the device and the currents of all superpixels of the NB classifier are recorded with an oscilloscope. The channel corresponding to the correct digit produces the highest output current. **b**, Experimental (symbols) and calculated (lines) light-intensity dependent accuracies for the NB classifier (blue) and a reference device without binning (red).

We conclude with proposed routes for future research. The main limitation of our current device implementation is its lack of reconfigurability. While this may be appropriate in some cases (e.g. a dedicated spectroscopic application), reconfigurability of the sensor would in general be preferred. This may, for example, be achieved by employing photodetectors with tunable responsivities, or a programmable network based on a nonvolatile memory material[23,24,25] to bin individual pixels together. Other schemes than standard one-hot encoding may allow to save hardware resources and extend the dynamic range further.

# METHODS

**Device fabrication.** Device fabrication started with the growth of a 400 nm thick $n^-$-doped ($10^{16}$ cm$^{-3}$) GaAs epilayer by molecular beam epitaxy on a highly $n^+$-doped GaAs substrate. An ohmic contact on the $n^+$-side was defined by evaporation of Ge/Au/Ni/Au (15 nm/30 nm/14 nm/300 nm) and sample heating at 440 °C for 30 s. On the $n^-$-GaAs epilayer we deposited a 20 nm thick Al$_2$O$_3$ insulating layer by atomic layer deposition (ALD). We then defined a first metal layer (M1) by electron-beam lithography (EBL) and Ti/Au (3 nm/25 nm) evaporation. In the next step we deposited a 30 nm thick Al$_2$O$_3$ layer by ALD. We then defined an etch mask for the via holes, which connect metal layers M1 and M2, by EBL and etched the Al$_2$O$_3$ with 30% potassium hydroxide (KOH) aqueous solution. We then wrote an etch mask for the pixel windows via EBL and etched the aggregated 50 nm thick Al$_2$O$_3$ with a 30% KOH aqueous solution in two steps. Inside the pixel windows, we defined the subpixels with EBL by removing the naturally formed oxide on the GaAs substrate with a 37% hydrochloric acid (HCl) aqueous solution and evaporating 7 nm thick semitransparent Au. Finally, we defined the M2 metal layer with EBL and Ti/Au (5 nm/80 nm) evaporation. The continuity and solidity of the device was confirmed by scanning electron microscopy and electrical measurements.

**Experimental setup.** A schematic of the experimental setup is shown in Figure 2b. A light-emitting diode (LED) source (625 nm wavelength) illuminates, through a linear polarizer, a spatial light modulator (SLM). The SLM is operated in intensity-modulation mode and changes the polarization of the reflected light according to the displayed image. The reflected light is then filtered using a second linear polarizer, and the image is projected onto the chip. The photocurrents generated by the sensor are probed with a needle array, selected by a Keithley switch matrix and measured with a Keithley source measurement unit. For time-resolved measurements a pulsed laser source (522 nm wavelength, 40 ns) is used. Here, the output signals are amplified with a high-bandwidth (20 MHz) transimpedance amplifier. The pulsed laser source is triggered with a signal generator and an oscilloscope is used to record the time trace.

**Noise measurements and modelling.** To characterize the noise performance, we performed binary image classification (NB, MNIST, '0' versus '1') at different light intensities. For the reference measurements, we projected the images sequentially, pixel by pixel, onto a single

GaAs Schottky photodetector (fabricated on the same wafer and with an area identical to that of two subpixels), recorded the photocurrents, and performed the classification task in a computer. In the simulations, Gaussian noise was added by drawing random samples from a normal distribution $\mathcal{N}(0, \sigma^2)$ with zero mean value. The noise was added once per superpixel in the data-driven case, and per each pixel in the reference case. $\sigma$ was used as a single fitting parameter to reproduce all experimental results.

**Data availability.** The data that support the findings of this study are available from the corresponding author upon reasonable request.

**Acknowledgments:** We thank Werner Schrenk, Fabian Dona and Andreas Kleinl for technical assistance. We acknowledge financial support by the Austrian Science Fund FWF (START Y 539-N16).

**Author contributions:** T.M. conceived the experiment. L.M. and T.M. designed the image sensor. L.M. built the experimental setup, programmed the machine learning algorithms, carried out the measurements, and analyzed the data. L.M., A.J.M.-M, M.P., D.K.P. and D.K. fabricated the device. M.G., M.B. and A.M.A. grew the GaAs wafer. T.M. and L.M. prepared the manuscript. All authors discussed the results and commented on the manuscript.

**Competing financial interests:** The authors declare no competing financial interests.

**REFERENCES**

[1] Boyle, W.S & Smith, G.E. Charge coupled semiconductor devices. *Bell Syst. Tech. J.* **49**, 587–593 (1970).

[2] El Gamal, A. & Eltoukhy, H. CMOS image sensors. *IEEE Circuits Devices Mag.* **21**, 6–20 (2005).

[3] Gallego, G. *et al.* Event-based vision: A survey. *IEEE Trans. Pattern Anal. Mach. Intell.*, DOI: 10.1109/TPAMI.2020.3008413 (2020).

[4] Posch, C., Serrano-Gotarredona, T., Linares-Barranco, B. & Delbruck, T. Retinomorphic event-based vision sensors: bioinspired cameras with spiking output. *Proc. IEEE* **102**, 1470–1484 (2014).

[5] Liao, F., Zhou, F. & Chai, Y. Neuromorphic vision sensors: principle, progress and perspectives. *J. Semicond.* **42**, 013105 (2021).

[6] Song, Y. M. *et al.* Digital cameras with designs inspired by the arthropod eye. *Nature* **497**, 95–99 (2013).

[7] Choi, C. *et al.* Human eye-inspired soft optoelectronic device using high-density MoS2-graphene curved image sensor array. *Nat. Commun.* **8**, 1664 (2017).

[8] Gao, S. *et al.* An oxide Schottky junction artificial optoelectronic synapse. *ACS Nano* **13**, 2634 (2019).

[9] Wang, H. *et al.* A ferroelectric/electrochemical modulated organic synapse for ultraflexible, artificial visual-perception system. *Adv. Mater.* **30**, 1803961 (2018).

[10] Seo, S. *et al.* Artificial optic-neural synapse for colored and color-mixed pattern recognition. *Nat. Commun.* **19**, 5106 (2018).

[11] Zhou, F. *et al.* Optoelectronic resistive random access memory for neuromorphic vision sensors. *Nat. Nanotechnol.* **14**, 776–782 (2019).

[12] Mennel, L., Symonowicz, J., Wachter, S., Polyushkin, D.K., Molina-Mendoza, A.J. & Mueller, T. Ultrafast machine vision with 2D material neural network image sensors. *Nature* **579**, 62–66 (2020).

[13] Wang, C.-Y. *et al.* Gate-tunable van der Waals heterostructure for reconfigurable neural network vision sensor. *Sci. Adv.* **6**, eaba6173 (2020).

[14] Jang, H. *et al.* An atomically thin optoelectronic machine vision processor. *Adv. Mater.* **32**, 2002431 (2020).

[15] Zhu, Q.-B. *et al.* A flexible ultrasensitive optoelectronic sensor array for neuromorphic vision systems. *Nat. Commun.* **12**, 1798 (2021).

[16] Chen, S., Lou, Z., Chen, D., Shen, G. An artificial flexible visual memory system based on an UV-motivated memristor. *Adv. Mater.* **30**, 1705400 (2018).

[17] Wang, S. *et al.* Networking retinomorphic sensor with memristive crossbar for brain-inspired visual perception. *Natl. Sci. Rev.* **8**, nwaa172 (2021).

[18] Epperson, P.M. & Denton, M.B. Binning spectral images in a charge-coupled device. *Anal. Chem.* **61**, 1513–1519 (1989).

[19] Lecun, Y., Bottou, L., Bengio, Y. & Haffner, P. Gradient-based learning applied to document recognition. *Proc. IEEE* **86**, 2278–2324 (1998).

[20] Bishop, C.M. *Pattern Recognition and Machine Learning* (Springer, 2007).

[21] Rennie, J. D., Shih, L., Teevan, J. & Karger, D. R. Tackling the poor assumptions of naive Bayes text classifiers. Proc. 20th Int. Conf. Machine Learning (ICML-03), pp. 616–623 (2003).

[22] Lazzaro, J., Ryckebusch, S., Mahowald, M. A. & Mead, C. A. Winner-Take-All Networks of O(N) Complexity. *Adv. Neural Inf. Process. Syst.* **1**, 703–711 (1989).

[23] Waser, R., Dittmann, R., Staikov, G. & Szot, K. Redox-based resistive switching memories - Nanoionic mechanisms, prospects, and challenges. *Adv. Mater.* **21**, 2632–2663 (2009).

[24] Yang, J.J., Strukov, D.B. & Stewart, D.R. Memristive devices for computing. *Nat. Nanotechnol.* **8**, 13–24 (2013).

[25] Burr, G.W. *et al.* Recent progress in phase-change memory technology. *IEEE J. Emerg. Sel. Topics Power Electron.* **6**, 146–162 (2016).